\documentclass[letterpaper, 10 pt, journal, twoside]{IEEEtran}

\IEEEoverridecommandlockouts                              

\usepackage{graphics} 
\usepackage{epsfig} 
\usepackage{times} 
\usepackage{amsmath} 
\usepackage{amssymb}  
\usepackage{hyperref} 
\usepackage{xcolor} 
\usepackage{arydshln} 
\usepackage{color} 
\usepackage{float}
\usepackage{multirow}
\usepackage{booktabs}
\usepackage{tikz}
\usepackage{wrapfig}
\usepackage{units}
\usepackage{bm}
\usetikzlibrary{calc ,math, positioning, intersections, decorations, external, plotmarks}
\usepackage{pgfplots}
\pgfplotsset{compat=1.16}
\usepackage[skip=0pt,font=footnotesize]{caption}
\usepackage{subcaption}

\newcommand{\rebuttal}[1]{{\color{black} #1}}
\newcommand\norm[1]{\left\lVert#1\right\rVert}

\title{
Learned Inertial Odometry for \\Autonomous Drone Racing
}

\author{Giovanni Cioffi, Leonard Bauersfeld, 
Elia Kaufmann, and Davide Scaramuzza
\thanks{Manuscript received: October, 27, 2022; Revised January, 20, 2023; Accepted February, 15, 2023.}
\thanks{
This paper was recommended for publication by Editor Pauline Pounds upon evaluation of the Associate Editor and Reviewers' comments.}
\thanks{This work was supported by the National Centre of Competence in Research (NCCR) Robotics through the Swiss National Science Foundation (SNSF) and the European Union’s Horizon 2020 Research and Innovation Programme under grant agreement No. 871479 (AERIAL-CORE) and the European Research Council (ERC) under grant agreement No. 864042 (AGILEFLIGHT).
}%
\thanks{The authors are with the Robotics and Perception Group, Department of Informatics, University of Zurich, and Department of Neuroinformatics, University of Zurich and ETH Zurich, Switzerland (\protect\url{http://rpg.ifi.uzh.ch}).}%
\thanks{Digital Object Identifier (DOI): see top of this page.}
}

\begin{document}
\makeatletter
\g@addto@macro\@maketitle{
  \captionsetup{type=figure}\setcounter{figure}{0}
  \def\mycolspace{1.2mm}
  \centering
    \includegraphics[width=1.75\columnwidth]{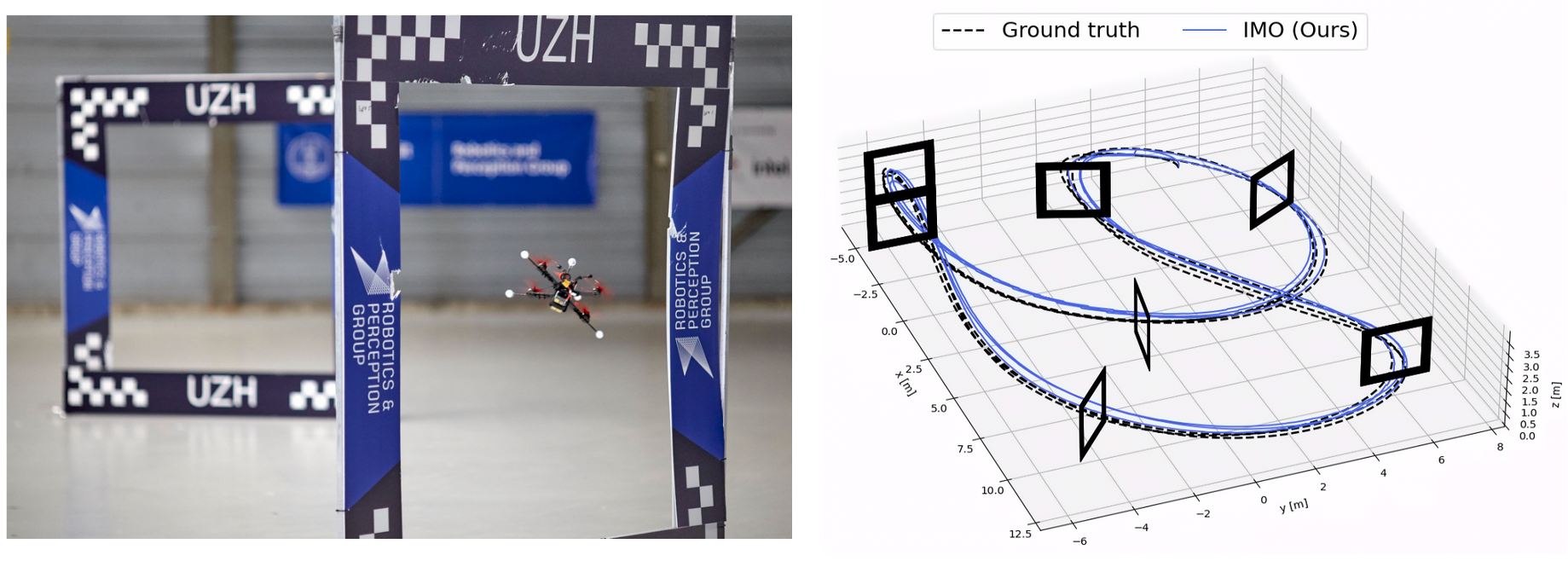}
	\captionof{figure}{By combining a temporal convolutional network in a model-based filter, our method is able to estimate the trajectory of an autonomous racing drone using an IMU as the only sensor modality. \textbf{Left}: Our autonomous drone flying in a race up to 70 $\frac{km}{h}$. \textbf{Right}: \rebuttal{The trajectory estimated by our method}.}
	\label{fig:eyecatcher}
}
\makeatother
\maketitle


\begin{abstract}

Inertial odometry is an attractive solution to the problem of state estimation for agile quadrotor flight.
It is inexpensive, lightweight, and it is not affected by perceptual degradation.
However, only relying on the integration of the inertial measurements for state estimation is infeasible.
The errors and time-varying biases present in such measurements cause the accumulation of large drift in the pose estimates.
Recently, inertial odometry has made significant progress in estimating the motion of pedestrians.
State-of-the-art algorithms rely on learning a motion prior that is typical of humans but cannot be transferred to drones.
In this work, we propose a learning-based odometry algorithm that uses an inertial measurement unit (IMU) as the only sensor modality for autonomous drone racing tasks.
The core idea of our system is to couple a model-based filter, driven by the inertial measurements, with a learning-based module that has access to the \rebuttal{thrust measurements}.
We show that our inertial odometry algorithm is superior to the state-of-the-art filter-based and optimization-based visual-inertial odometry as well as the state-of-the-art learned-inertial odometry in estimating the pose of an autonomous racing drone.
Additionally, we show that our system is comparable to a visual-inertial odometry solution that uses a camera and exploits the known gate location and appearance.
We believe that the application in autonomous drone racing paves the way for novel research in inertial odometry for agile quadrotor flight.
\end{abstract}

\begin{IEEEkeywords}
Aerial Systems: Perception and Autonomy; Aerial Systems: Applications; Deep Learning Methods
\end{IEEEkeywords}

\section*{Supplementary Material}\label{sec:SupplementaryMaterial}

\textbf{Video}:\url{https://youtu.be/2z2Slyt0WlE}

\textbf{Code}:\url{https://github.com/uzh-rpg/learned_inertial_model_odometry}
\section{Introduction}\label{sec:Introduction}


\IEEEPARstart{Q}{uadrotors} are extremely agile.
Making them autonomous is crucial for time-critical missions, such as search and rescue, aerial delivery, reconnaissance, and even flying cars~\cite{loianno2020special,watkins2020ten, rakha2018review}.
State estimation is a core block of the autonomy pipeline.

Inertial odometry (IO) is an excellent solution to the problem of state estimation for agile quadrotor flight.
Inertial Measurements Units (IMUs) are inexpensive and ubiquitous sensors that provide linear accelerations and angular velocities.
An odometry algorithm only based on inertial measurements has low power and storage requirements, and it does not suffer in scenarios where vision-based odometry systems commonly fail, e.g. large motion blur, high dynamic range scenes, and low texture environments.
These are the typical scenarios encountered in agile quadrotor flight.
In theory, inertial measurements can be integrated to obtain 6-DoF poses.
In practice, the measurements provided by off-the-shelf IMUs are affected by scale factor errors, axis misalignment errors, and time-varying biases~\cite{yang2020online}.
Consequently, the integration accumulates large drift in a short time.

Recently, major progress has been made in inertial odometry for state estimation of pedestrian motion~\cite{chen2018ionet, herath2020ronin, liu2020tlio}.
These works have shown that motion priors can be learned from the repetitive pattern of human gait using IMU measurements.
The accuracy of these IO algorithms is comparable to the one of visual-inertial odometry (VIO) algorithms for pedestrian applications.

Differently from the pedestrian motion, the quadrotor motion is not characterized by any significant prior that can be learned from the IMU measurements.
For this reason, the performance of the IO methods proposed for pedestrian navigation deteriorates when applied to quadrotors.
In this work, we propose a learned inertial odometry algorithm to tackle the problem of state estimation in autonomous drone racing.

Why drone racing?
Drone racing requires flying a drone through a sequence of gates in minimum time and has now become a benchmark for the development of new drone technologies that can be turned into real-world products~\cite{madaan2020airsim, foehn2021alphapilot}.
What makes drone racing so challenging is that the platform is flown at incredible speeds, close to a hundred kilometers per hour, pushing the boundaries of the physics of the vehicle. At such speeds, any little state estimation error can lead to a crash.
Research works so far mainly focused on planning and control~\cite{foehn2021time, romero2022model, penicka2022minimum}, and relied on external position-tracking systems for state estimation.
However, tackling the state estimation problem with only onboard sensing is key to achieving full autonomy.
In absence of external position-tracking systems, the only viable solution for state estimation of flying vehicles is VIO~\cite{zhang2019encyclopedia,huang2019visual}. 
However, VIO fails in scenarios characterized by motion blur, low texture, and high dynamic range due to the unreliability of the vision system. 
These failure cases are always present in drone racing.
Conversely, inertial odometry is not affected by these challenges.

We propose a learned inertial odometry algorithm that uses an IMU as the only sensor modality. 
Our algorithm combines an Extended Kalman Filter~(EKF), which is driven by the inertial measurements, with a learning-based module, which has access to \rebuttal{measurements that are related to the drone dynamics} in the form of mass-normalized collective thrust.
The learning-based module is a temporal convolutional network~(TCN) that takes as input a buffer of mass-normalized collective thrust and gyroscope measurements and outputs an estimate of the distance traveled by the quadrotor.
These relative positional displacements are then used to update the filter.

We show that the proposed algorithm is superior to the state-of-the-art filter-based and optimization-based VIO algorithms as well as the state-of-the-art learned inertial odometry algorithm~TLIO~\cite{liu2020tlio} in estimating the pose of a racing quadrotor.
Additionally, our approach achieves comparable results to a VIO solution that uses a camera and exploits the known gate location and appearance.
We believe that the application in autonomous drone racing shows the benefits of our method and paves the way for novel research in inertial odometry for agile quadrotor flight.

The main contributions of this work are:
\begin{itemize}
    \item An inertial odometry algorithm based on an EKF that is propagated by the inertial measurements and is updated by the relative positional displacements predicted by a temporal convolutional network.
    We name our algorithm \textit{IMU-Model Odometry (IMO)}.
    \item \rebuttal{Validation of the proposed system in multiple agile quadrotor flights from the Blackbird dataset~\cite{Antonini18iser}.}
    \item Thorough experimental analysis for drone racing. Our analysis includes comparisons of the proposed approach against the state-of-the-art VIO and learned IO algorithms as well as against a VIO algorithm that uses a camera to localize to the gates.
    \item Ablation studies validate both the model-based component, EKF, and the learning-based component, TCN, of our system.
\end{itemize}
\section{Related Work}\label{sec:RelatedWork}

The most common solution for state estimation of aerial vehicles using only onboard sensing is VIO thanks to its low power, low cost, and low weight requirements.
VIO algorithms are divided into two categories according to how they fuse the camera and IMU measurements: filtering methods~\cite{mourikis2007multi} and fixed-lag smoothing methods~\cite{Leutenegger15ijrr}.
Filtering methods are based on the EKF.
These methods propagate the state of the system using the IMU measurements and fuse the camera measurements in the update step.
Filtering methods achieve a favorable trade-off between computational requirements and accuracy.
Fixed-lag smoothing methods solve a non-linear optimization problem where camera reprojection, IMU, and marginalization residuals are optimized in a sliding window  fashion.
These methods accumulate less linearization error than filter-based approaches but they are more computationally expensive.
We refer the reader to the survey work in~\cite{huang2019visual} for more details.

Recent works have shown relevant progress in inertial odometry for pedestrian motion estimation by combining model-based approaches with deep learning~\cite{chen2018ionet, herath2020ronin, liu2020tlio}.
The works in~\cite{chen2018ionet, herath2020ronin} have shown that 2-DoF pedestrian trajectories can be accurately estimated using neural networks that are trained to predict velocities using IMU data collected from hand-held devices.
These works are extended by~\cite{liu2020tlio}, where it is proposed a 1-D residual convolutional network architecture that learns relative position displacements using IMU data collected from a VR headset device.
These relative position displacements are then used as measurements to update an EKF.
In robotic applications, deep learning approaches have been used to denoise gyro measurements that are afterward integrated for attitude estimation~\cite{brossard2020denoising}, to denoise IMU measurements before they are included in a VIO algorithm~\cite{zhang2021imu}, and to compute IMU factors in a sensor fusion algorithm~\cite{buchanan2021learning}.

The works in~\cite{nisar2019vimo, ding2021vid} propose using the quadrotor dynamics in VIO. The commanded collective thrust is integrated to derive pre-integrated positional factors that are included in an optimization-based VIO algorithm.
The difference between these two works is in the stochastic model of the external force.
In~\cite{zhang2022dido}, the authors propose a system that estimates the full state of a quadrotor using an IMU and 4 tachometers attached to the rotors.
Their approach relies on a heavy recurrent network that is trained to minimize the drift of the full trajectory.
The main difference between our work and~\cite{zhang2022dido} is that our method uses a lightweight network, which can run on the computer onboard the drone, and does not rely on rotor speed measurements, which are often not available in real-time onboard drone platforms.
Another work that relies on rotor speed measurements, as well as inertial measurements, is the one in~\cite{svacha2020imu}. 
This method estimates the tilt, linear and angular velocities as well as additional model parameters, such as the moment of inertia, of the quadrotor but its position.
\section{Methodology}\label{sec:Methodology}

\begin{figure}[t!]
    \centering
    \includegraphics[width=1.0\linewidth]{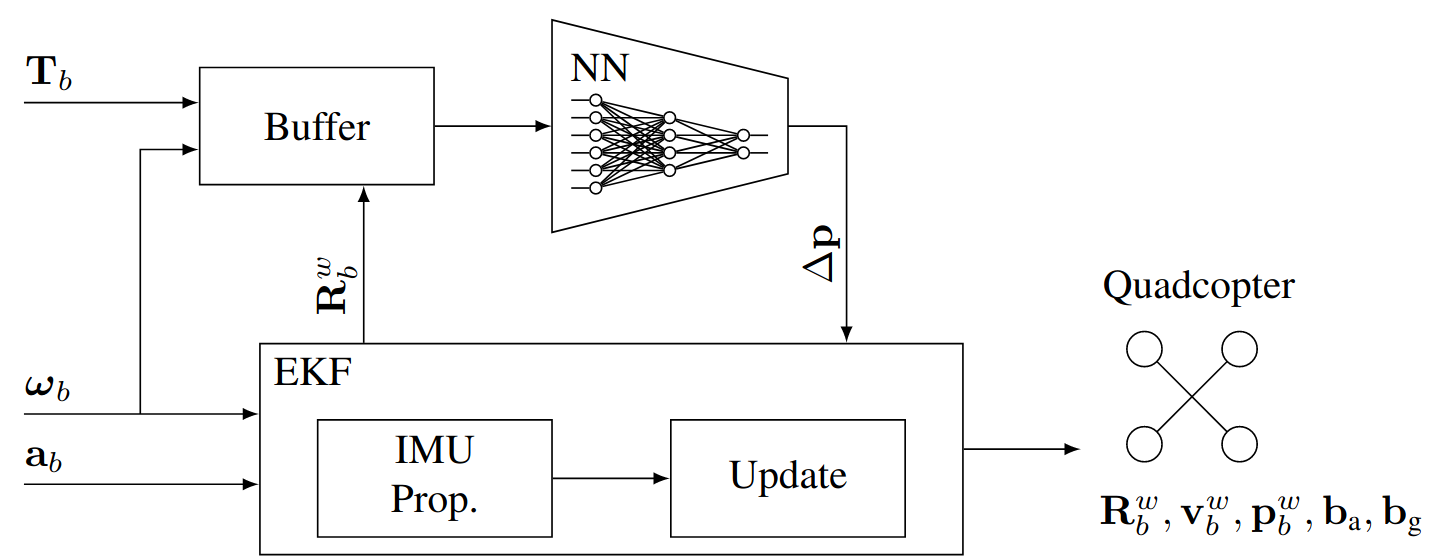}
    \caption{Block diagram of our system. A neural network takes as input a buffer of collective thrust and gyroscope measurements and outputs relative 3-DoF positional displacements. These displacements are used to update an EKF, which is propagated using the IMU measurements.}
    \label{fig:methodology_flowchart}
\end{figure}

An overview of our system is shown in Fig.~\ref{fig:methodology_flowchart}.
We train a temporal convolutional network~\cite{oord2016wavenet} to regress 3-DoF relative displacements from a buffer of length $\Delta t$ seconds containing mass-normalized collective thrust and gyroscope measurements.
Collective thrust has been shown to be the preferred choice of control inputs for agile quadrotor flight~\cite{kaufmann2022benchmark}.
These relative displacements represent the distance traveled by the quadrotor in the time interval $\Delta t$.

We train the neural net in order to learn a prior on the translational motion of the quadrotor.
The displacements predicted by the neural net are used as measurements to update an EKF.
The EKF is propagated using a kinematic motion model of the IMU.

In this section, first, we introduce the notation used throughout the paper.
Second, we describe how we leverage deep learning in the quadrotor model.
Third, we describe the EKF.
Fourth, we introduce Gate-IO, a VIO algorithm developed for the task of state estimation in drone racing.
We use Gate-IO as a baseline to validate the performance of the proposed system.
Last, we describe the main implementation details.

\subsection{Notation}\label{sec:notation}

The reference frame $\mathcal{W}$ is the fixed world frame, whose $z_w$ axis is aligned with gravity.
The quadrotor body frame is $\mathcal{B}$. 
For simplicity, the IMU frame is assumed to be the same as $\mathcal{B}$.
We use the notation $(\cdot)^{w}$ to represent a quantity in the world frame $\mathcal{W}$. 
A similar notation applies to each reference frame.
The position, orientation, and velocity of $\mathcal{B}$ with respect to $\mathcal{W}$ at time $t_k$ are written as $\mathbf{p}_{b_{k}}^{w} \in \mathbb{R}^3$, $\mathbf{R}_{b_{k}}^{w} \in \mathbb{R}^{3 \times 3}$ part of the rotation group $SO(3)$, and $\mathbf{v}_{b_{k}}^{w} \in \mathbb{R}^3$, respectively.
The accelerometer and gyroscope bias are written as $\mathbf{b}_{a}$ and $\mathbf{b}_{g}$, respectively.
The gravity vector in the world frame is written as $\mathbf{g}^w$.
We denote estimated quantities with $\hat{(\cdot)}$ and measured quantities with $\tilde{(\cdot)}$.

\subsection{Learned Quadrotor Model}\label{sec:learned_quadrotor_model}
\subsubsection{Quadrotor Model}\label{sec:learned_quadrotor_model}

The evolution of the position and velocity of the quadrotor platform is described by the following model~\cite{nisar2019vimo}:
%
\begin{equation}\label{eq:drone_model}
    \dot{\mathbf{p}}^{w}_{b_i} = \mathbf{v}^{w}_{b_i},\;\;
    \dot{\mathbf{v}}^{w}_{b_i} = \mathbf{R}^{w}_{b_i} \cdot (\mathbf{T}^{b}_{i} + \mathbf{F}^{b}_{e_i}) + \mathbf{g}^{w},
\end{equation}
where $\mathbf{T}^{b}_{i}$ is the mass-normalized collective thrust and $\mathbf{F}^{b}_{e_i}$ is the external force acting on the platform.
We will drop the term mass-normalized when referring to the collective thrust hereafter for the sake of conciseness. 
Since we do not know the dynamics of the external force, we assume it to be a random variable distributed according to a zero-mean Gaussian distribution~\cite{nisar2019vimo}.
Integrating Eq.~\ref{eq:drone_model} in the time interval $[t_i, t_{i+1}]$ with the assumption that $\mathbf{T}^{b}_{i}$ is constant in such an interval and using $\mathbf{F}^{b}_{e_i} = [0, 0, 0]$, we obtain an explicit relation between the relative positional displacement and the thrust:
\begin{equation}\label{eq:dp}
    \Delta \mathbf{p}_{i,i+1} = \mathbf{v}^{w}_{b_i} \cdot \Delta t + 0.5 \cdot \mathbf{g}^{w} \cdot \Delta t^2 + 0.5 \cdot \mathbf{R}^{w}_{b_i} \cdot \mathbf{T}^{b}_{i} \cdot \Delta t^2,
\end{equation}
and
\begin{equation}\label{eq:thrust}
    \mathbf{T}^{b}_{i} = 2 \cdot \mathbf{R}_{w}^{b_i} \left(\frac{\Delta \mathbf{p}_{i,i+1}}{\Delta t^2} - \frac{\mathbf{v}^{w}_{i}}{\Delta t} - \mathbf{g}^{w}\right),
\end{equation}
where $\Delta \mathbf{p}_{i,i+1} = \mathbf{p}^{w}_{b_{i+1}} - \mathbf{p}^{w}_{b_i}$.

\subsubsection{Neural Net Model}\label{sec:network_architecture}

In this work, we use a TCN to learn the positional displacements $\Delta \mathbf{p}_{i,j}$.
TCNs have been shown to be as powerful as recurrent networks to model temporal sequences~\cite{bai2018empirical} but they are easier to train and deploy on a robotic platform.
The neural net takes as input a buffer of collective thrust and gyroscope measurements.
These measurements are rotated to the world frame and the bias is removed from the gyroscope measurements.
During training, we use ground-truth orientations obtained from a motion capture system.
At deployment time, we use the orientations estimated by the EKF.
To increase the robustness of the neural net to uncertainty in the estimated orientation, we perturb the ground-truth orientations at training time with zero-mean Gaussian noise.
The standard deviation of this noise depends on the expected accuracy of the orientations estimated by the filter.
We apply the same strategy to increase the robustness of the neural net with respect to uncertainty in the estimate of the gyroscope bias.
Given as input a buffer of measurements in the time interval $\Delta t_{i,j} = t_j - t_i$, the neural net output is the relative displacement $\Delta \mathbf{p}_{i,j}$ in $\Delta t_{i,j}$.
We train the neural net with the MSE loss:
\begin{equation}\label{equ:net_loss}
    \mathcal{L}(\Delta \mathbf{p}, \Delta \hat{\mathbf{p}}) = \frac{1}{N} \sum_{j = 1}^{N} \norm{\Delta \mathbf{p}_k - \Delta \hat{\mathbf{p}}_k}^{2}
\end{equation}
where $\Delta \mathbf{p}$ is the ground-truth positional displacement. 
We omitted the temporal indices $i,j$ in Eq.~\ref{equ:net_loss} for the sake of conciseness.

\subsection{Inertial Model Odometry}\label{sec:imo}

In this section, we describe the core components of the EKF.\\
\textbf{Filter state}. Our EKF is based on~\cite{mourikis2007multi}.
The full state of the filter is \rebuttal{$\mathcal{X} = \{\zeta_1, \cdots , \zeta_m,  s\}$}, where $s = \{\hat{\mathbf{R}}_{b_{i}}^{w}, \hat{\mathbf{v}}_{b_{i}}^{w}, \hat{\mathbf{p}}_{b_{i}}^{w}, \hat{\mathbf{b}}_{a_{i}}, \hat{\mathbf{b}}_{g_{i}}\}$ is the current filter state and $\zeta_j =\{\hat{\mathbf{R}}_{b_{j}}^{w}, \hat{\mathbf{p}}_{b_{j}}^{w}\} $ is the $j$-th past state.
Following the error-based filtering
approach~\cite{mourikis2007multi}, we linearize on a manifold of minimal parameterization of the rotation.
The error-state representation of the current filter state is $\delta \mathbf{s} = \{ \delta \mathbf{\theta}_{b_{i}}^{w}, \delta \mathbf{v}_{b_{i}}^{w}, \delta \mathbf{p}_{b_{i}}^{w}, \delta \mathbf{b}_{a_{i}}, \delta \mathbf{b}_{g_{i}} \}$.
For an arbitrary variable $\mathbf{x}$, we define $\delta \mathbf{x} = \mathbf{x} \boxminus \hat{\mathbf{x}}$.
The operator $\boxminus$ is the difference between the ground-truth and the estimated value. 
This is the difference operator for variables $\in \mathbb{R}^3$ and the 
logarithm map operator, $\text{Log}(\cdot)$, such that
$\mathbf{\theta} = \text{Log}(\mathbf{R} \cdot \hat{\mathbf{R}}^{-1})$ for $\mathbf{R} \in SO(3)$. 
Its inverse is the exponential map $\text{Exp}(\cdot)$.\\
\textbf{Filter propagation}. We use the IMU model as in~\cite{mourikis2007multi}. The gyroscope model is: $\tilde{\bm{\omega}}_i = \bm{\omega}_i + \mathbf{b}_g + \mathbf{n}_g$, and the accelerometer is: $\tilde{\mathbf{a}}_i = \mathbf{a}_i + \mathbf{b}_{a} + \mathbf{n}_{a}$.
This model represents the measurements as the ground-truth values corrupted by bias and zero-mean Gaussian noise.
The accelerometer and gyroscope biases are modeled as a random walk processes with noise $\mathbf{n}_{\text{b}_{a}}$ and $\mathbf{n}_{\text{b}_{g}}$, respectively.
The kinematic motion model of the filter propagation is
\begin{align}\label{equ:ekf_propagation}
\hat{\mathbf{R}}_{b_{i+1}}^{w} &= \hat{\mathbf{R}}_{b_{i}}^{w} \cdot \text{Exp}(\tilde{\bm{\omega}}_i - \hat{\mathbf{b}}_{g_{i}}) \cdot \Delta t \nonumber \\
\hat{\mathbf{v}}_{b_{i+1}}^{w} &= \hat{\mathbf{v}}_{b_{i}}^{w} + \mathbf{g}^{w} \cdot \Delta t + \hat{\mathbf{R}}_{b_{i}}^{w} \cdot (\tilde{\mathbf{a}}_i - \hat{\mathbf{b}}_{a_i}) \cdot \Delta t \nonumber \\
\hat{\mathbf{p}}_{b_{i+1}}^{w} &= \hat{\mathbf{p}}_{b_{i}}^{w} + \hat{\mathbf{v}}_{b_{i}}^{w} \cdot \Delta t + \frac{1}{2} \cdot \Delta t^2 \cdot (\mathbf{g}^{w} + \hat{\mathbf{R}}_{b_{i}}^{w} \cdot (\tilde{\mathbf{a}}_i - \hat{\mathbf{b}}_{a_i}))\nonumber \\
\hat{\mathbf{b}}_{g_{i+1}} &= \hat{\mathbf{b}}_{g_i}, \space
\hat{\mathbf{b}}_{a_{i+1}} = \hat{\mathbf{b}}_{a_i}.
\end{align}
To simplify the notation, we dropped the indices that refer to the value of the state between the propagation and update steps and wrote $\Delta t = \Delta t_{i,i+1}$.
The linearized propagation model is
\begin{equation}\label{equ:error_state_prop}
    \delta \mathbf{s}_{i+1} = \mathbf{A}_{i} \cdot \delta \mathbf{s}_i + \mathbf{B}_i \cdot \mathbf{n}_s,
\end{equation}
where $\mathbf{n}_{s}^{\intercal} = [\mathbf{n}_{a}^{\intercal}, \mathbf{n}_{g}^{\intercal}, \mathbf{n}_{\text{b}_{a}}^{\intercal}, \mathbf{n}_{\text{b}_{g}}^{\intercal} ]$ is the vector containing the IMU noise and IMU biases noise. 
The covariance propagation model is
\begin{equation}\label{equ:cov_prop}
    \mathbf{P}_{i+1} = \mathbf{A}^{\mathcal{X}}_{i} \cdot \mathbf{P}_{i} \cdot (\mathbf{A}^{\mathcal{X}}_{i})^{t} + \mathbf{B}^{\mathcal{X}}_{i} \cdot \mathbf{W}_{i} \cdot (\mathbf{B}^{\mathcal{X}}_{i})^{t},
\end{equation}
where $\mathbf{W}_i$ is the covariance matrix containing the IMU noise and IMU bias noise and
\begin{align}
    \mathbf{A}^{\mathcal{X}}_i = \begin{bmatrix} \mathbf{I} & \mathbf{0} \\ \mathbf{0} & \mathbf{A}_i \end{bmatrix},
    \mathbf{B}^{\mathcal{X}}_i = \begin{bmatrix} \mathbf{0} \\ \mathbf{B}_i \end{bmatrix}.
\end{align}
The matrix $\mathbf{I}$ denotes the identity matrix whose dimensions depend on the number of past states.\\
\textbf{State augmentation}. 
The state augmentation is performed at the filter update frequency.
The full state is augmented with a new state that is obtained by propagation till the time when a new measurement from the TCN is available.
The covariance propagation is similar as in Eq.~\ref{equ:cov_prop} with
\begin{align}
    \mathbf{A}^{\mathcal{X}}_i = \begin{bmatrix} \mathbf{I} & \mathbf{0} \\ \mathbf{0} & \mathbf{A}^{\zeta}_i \\ \mathbf{0} & \mathbf{A}_i \end{bmatrix},
    \mathbf{B}^{\mathcal{X}}_i = \begin{bmatrix} \mathbf{0} \\ \mathbf{B}^{\zeta}_i \\ \mathbf{B}_i \end{bmatrix}.
\end{align}
%
%
\textbf{Filter update}. 
The measurement update is: $r(\mathcal{X}) = \hat{\mathbf{p}}^{w}_{j} - \hat{\mathbf{p}}^{w}_{i} = \Delta \tilde{\mathbf{p}}_{ij} + \mathbf{n}_{ij}$, where
$\mathbf{n}_{ij}$ is a zero-mean Gaussian noise on the predictions of the network.
We set it as a constant value.
The Jacobian matrix $\mathbf{H}$ is straightforward to compute.
Its entries are all zero except for
\begin{equation}
    \mathbf{H}_{\hat{\mathbf{p}}_i^w} = \frac{\partial r(\mathcal{X}) }{ \partial \delta \hat{\mathbf{p}}_i^w} = - \mathbf{I}_{3},
    \mathbf{H}_{\hat{\mathbf{p}}_j^w} = \frac{\partial r(\mathcal{X}) }{ \partial \delta \hat{\mathbf{p}}_j^w} =  \mathbf{I}_{3},
\end{equation}
where $\mathbf{I}_{3}$ is the identity matrix with dimensions $3 \times 3$.
The filter update is performed as
\begin{align}\label{eq:kalman_gain_and_update}
    \mathbf{K} &= \mathbf{P} \cdot \mathbf{H}^t \cdot (\mathbf{H} \cdot \mathbf{P} \cdot \mathbf{H}^t + \mathbf{\Sigma}_{ij})^{-1}\\
    \mathcal{X} &= \mathcal{X} \boxplus (\mathbf{K} \cdot (\hat{\mathbf{p}}_j^w - \hat{\mathbf{p}}_i^w - \Delta \tilde{\mathbf{p}}_{ij}))\\
    \mathbf{P} &= (\mathbf{I} - \mathbf{K} \cdot \mathbf{H}) \cdot \mathbf{P} \cdot (\mathbf{I} - \mathbf{K} \cdot \mathbf{H})^t + \mathbf{K} \cdot \mathbf{\Sigma}_{ij} \cdot \mathbf{K}^t.
\end{align}
The operator $\boxplus$ represents the addition for variables $\in \mathbb{R}^3$ and the update operation $\mathbf{R}^{\prime} = \text{Exp}(\theta) \cdot \mathbf{R}$ for variables $\in SO(3)$.

\subsection{Gate-IO}\label{sec:gateio}

In this work, we develop a VIO algorithm for the task of state estimation in autonomous drone racing. We name this algorithm: Gate-IO.
Gate-IO fuses the detection of the gate corners with IMU measurements in an EKF.
The EKF state and the propagation step are the same as the ones described in Sec.~\ref{sec:imo}.
The residual function of the update step is the reprojection of the gate corners onto the image frame.
The world coordinates of the gate corners are known beforehand.
The gate corners are detected using a convolutional neural network~(CNN) that takes as input raw camera images and outputs the pixel coordinates of all the visible gate corners.
The CNN is specifically trained for the type of gates used in our experiments.
This approach is based on~\cite{foehn2021alphapilot}.

\subsection{Implementation Details}\label{implementation_details}

We train our neural net on a laptop running Ubuntu 20.04 and equipped with an Intel Core i9 2.3GHz CPU and Nvidia RTX 4000 GPU.
At test time, our system runs on an NVIDIA Jetson TX2, which is the computing platform onboard the quadrotor.
All the baselines run on the laptop.
The thrust and gyroscope measurements are sampled at 100 Hz and are fed to the neural net in an input buffer of length 0.5 seconds.
The neural net inference, and consequently the EKF update frequency, is set to 20 Hz.
The maximum number of the past states in the filter state is 10.
With this setting, our system runs at $\sim$180 Hz on the Jetson TX2 onboard the quadrotor.
The noise values are: 
$\sigma_{\mathbf{n}_a} = 0.01 \frac{\text{m}}{\text{s}^2}, \sigma_{\mathbf{n}_g} = 0.001 \frac{\text{rad}}{\text{s}}, \sigma_{\mathbf{n}_{\mathbf{b}_a}} = 0.001 \frac{\text{m}}{\text{s}^2} \frac{1}{\text{s}^{0.5}}, \sigma_{\mathbf{n}_{\mathbf{b}_g}} = 0.0001 \frac{\text{rad}}{\text{s}} \frac{1}{\text{s}^{0.5}}, \sigma_{\mathbf{n}_{ij}} = 0.01 \text{m}$.
\rebuttal{The value of $\sigma_{\mathbf{n}_{ij}}$ has been chosen according to the expected error of the network predictions.}
We initialize our method as well as all the baselines with the ground-truth initial \rebuttal{pose} obtained from a motion caption system.
\rebuttal{In the experiments on the Blackbird dataset, the IMU biases are initialized to zero. In the drone racing experiments, the IMU biases are initialized using an offline calibration routine.}
\section{Experiments}\label{sec:Experiments}

We compare our system to the following baselines:
\begin{itemize}
    \item TLIO~\cite{liu2020tlio}. TLIO is the state-of-the-art inertial odometry algorithm. It uses a residual network that takes as input a buffer of IMU measurements in a time window and outputs the relative distance that the IMU has traveled in such a time window.
    The network output is used as a measurement to update an EKF. The EKF is propagated using the IMU measurements.
    \item SVO~\cite{Forster17troSVO}. SVO is a semi-direct visual odometry front-end. In this work, we combine SVO with a sliding-window optimization-based backend~\cite{Leutenegger15ijrr}. The code is available open-source~\footnote{\url{https://github.com/uzh-rpg/rpg_svo_pro_open}}.
    \item OpenVINS~\cite{geneva2020openvins}. OpenVINS is a state-of-the-art filter-based visual-inertial odometry algorithm. It ranks 1-st amongst the open-source algorithms on the UZH-FPV drone racing dataset~\cite{Delmerico19icra}.
    \item Gate-IO. Gate-IO is a VIO algorithm customized for drone racing. We refer the reader to Sec.~\ref{sec:gateio} for more details.
\end{itemize}
Following the best practices in the evaluation of VIO algorithms~\cite{Zhang18iros}, we use the evaluation metrics: translation absolute trajectory error ($\text{ATE}_{\text{T}}$) [m], rotation absolute trajectory error ($\text{ATE}_{\text{R}}$) [deg], relative translation and rotation errors. We refer the reader to~\cite{Zhang18iros} for a detailed description of these metrics.
%

\begin{figure}[t!]
    \centering
    \includegraphics[width=0.90\linewidth]{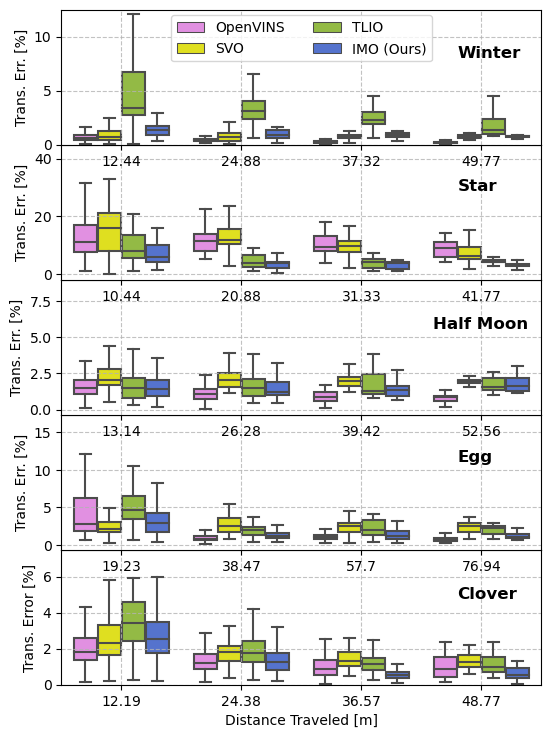}
    \caption{\rebuttal{Blackbird dataset evaluation. Relative translation errors achieved by OpenVINS, SVO, TLIO, and IMO (ours). The quantity on the x-axis is the distance traveled corresponding to 2.5, 5, 7.5, and 10 \% of the total distance traveled.}}
    \label{fig:blackbird_reltrans_errs}
\end{figure}

\begin{figure}[t!]
    \centering
    \includegraphics[width=0.9\linewidth]{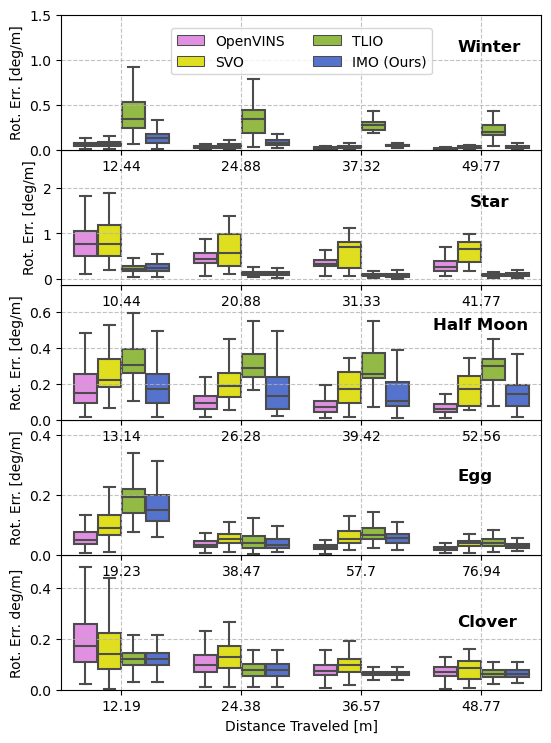}
    \caption{\rebuttal{Blackbird dataset evaluation. Relative rotation errors achieved by OpenVINS, SVO, TLIO, and IMO (ours). The quantity on the x-axis is the distance traveled corresponding to 2.5, 5, 7.5, and 10 \% of the total distance traveled.}}
    \label{fig:blackbird_relrot_errs}
\end{figure}

\rebuttal{
\subsection{Blackbird Dataset}\label{sec:experiments_on_blackbird}

\textit{Experiment Setup:} 
In this set of experiments, we evaluate the performance of our system and the baselines on the Blackbird dataset~\cite{Antonini18iser}.
The Blackbird dataset provides rotor speed measurements recorded onboard a quadrotor flying in a motion capture system, which we use to compute mass-normalized collective thrust measurements for our network.
In addition, this dataset also contains IMU measurements and photorealistic images of synthetic scenes.
We select 5 trajectories from the dataset: \textit{clover}, \textit{egg}, \textit{half moon}, \textit{star}, and \textit{winter}, with peak velocities of 5, 8, 4, 5, 4 $\frac{m}{s}$, respectively.
For each trajectory, $70\%$ of the data is used for training, $15\%$ of the data is used for validation and, $15\%$ of the data is used for testing.
In total, the training, validation, and test datasets contain approx. 10, 2.5, and 2.5 min of flight data, respectively.
We use the training and validation dataset to train our network and the TLIO network and to tune the parameters of SVO and OpenVINS.

\textit{Evaluation:}
We report the absolute trajectory errors in Table~\ref{tab:ate_blackbird} and the relative translation and rotation errors in Fig.~\ref{fig:blackbird_reltrans_errs} and Fig.~\ref{fig:blackbird_relrot_errs}, respectively.
Our system outperforms TLIO in all the sequences.
The smallest and the largest improvements of the $\text{ATE}_{\text{T}}$ are equal to $12\%$ and $80\%$ in the sequences \textit{egg} and \textit{winter}, respectively.
The best VIO algorithm is OpenVINS.
The performance of our system is comparable to the one of OpenVINS in all the sequences.
The largest difference in the $\text{ATE}_{\text{T}}$ in favor of our system is in the \textit{star} sequence.
In this sequence, the high yaw rate and the resulting large optical flow render feature tracking more difficult and, consequently, degrades the estimation accuracy of the VIO system.}
\begin{table}[t!]
\caption{
\rebuttal{Blackbird dataset evaluation. $\text{ATE}_{\text{T}}$ is in meters and $\text{ATE}_{\text{R}}$ is in degrees. In bold is the best value and in underlined is the second-best value.}}
\label{tab:ate_blackbird}
\resizebox{0.49\textwidth}{!}{
\begin{tabular}{|c|c|c|c|c|c|}
\hline
\multirow{2}{*}{\textbf{\begin{tabular}[c]{@{}l@{}}Trajectory\end{tabular}}} & \multirow{2}{*}{\textbf{\begin{tabular}[c]{@{}c@{}}Eval. \\ metric\end{tabular}}} & \multicolumn{4}{c|}{\textbf{Algorithm}} \\ \cline{3-6} 
 & & \textbf{OpenVINS} & \textbf{SVO} & \textbf{TLIO} & \textbf{IMO (ours)}\\ \hline \hline 

\multirow{2}{*}{\textit{Clover}} & $\text{ATE}_{\text{T}}$ [m] & \underline{0.50} & 0.77 & 0.75 & \textbf{0.41} \\ & $\text{ATE}_{\text{R}}$ [deg] & \textbf{2.62} & 3.51 & \underline{3.05} & \underline{3.05}\\
\hline
\multirow{2}{*}{\textit{Egg}} & $\text{ATE}_{\text{T}}$ [m] & \textbf{1.07} & 2.49 & 1.31 & \underline{1.15} \\ & $\text{ATE}_{\text{R}}$ [deg] & \underline{2.71} & 3.42 &  2.97 & \textbf{2.45}\\
\hline
\multirow{2}{*}{\textit{Half Moon}} & $\text{ATE}_{\text{T}}$ [m] & \textbf{0.37} & 1.10 & 1.20 & \underline{0.76} \\ & $\text{ATE}_{\text{R}}$ [deg] & \textbf{2.29} & 8.48 & 8.74 & \underline{4.14}\\
\hline
\multirow{2}{*}{\textit{Star}} & $\text{ATE}_{\text{T}}$ [m] & 2.78 & 2.78 & \underline{2.04} & \textbf{1.22} \\ & $\text{ATE}_{\text{R}}$ [deg] & 7.43 & 10.16 & \underline{2.96} & \textbf{2.76}\\
\hline
\multirow{2}{*}{\textit{Winter}} & $\text{ATE}_{\text{T}}$ [m] & \textbf{0.12} & 0.29 & 1.13 & \underline{0.22} \\ & $\text{ATE}_{\text{R}}$ [deg] & \textbf{0.87} & \underline{1.18} & 12.15 & 2.32\\
\hline
\end{tabular}
}
\end{table}

\subsection{Drone Racing}\label{sec:experiments_racing}

\textit{Experiment Setup:} 
To validate our system in drone racing tasks, we designed a custom-made quadrotor platform.
The platform has a total weight of 750 grams and it can produce a maximum thrust larger than 40 N.
The weight and power of this platform are comparable to the ones used by professional pilots in drone racing competitions.
The main computational unit is an NVIDIA Jetson TX2.
Our quadrotor is also equipped with an Intel RealSense T265 camera.
This camera has a fisheye lens and provides grey-scale images with a resolution of 848$\times$800 pixels.
The camera also contains an integrated IMU.
SVO, OpenVINS, and Gate-IO use the monocular grey-scale images and inertial measurements from the Intel RealSense T265 while TLIO only uses the inertial measurements.
More details about the quadrotor platform and the onboard software stack can be found in~\cite{foehn2022agilicious}.
In all our experiments, the quadrotor is flown in a motion capture system and controlled by the method proposed in~\cite{song2021autonomous}.
The controller~\cite{song2021autonomous} outputs collective thrusts. 
These commanded collective thrusts are used as input to our network.

\begin{figure}[t!]
    \centering
    \includegraphics[width=0.9\linewidth]{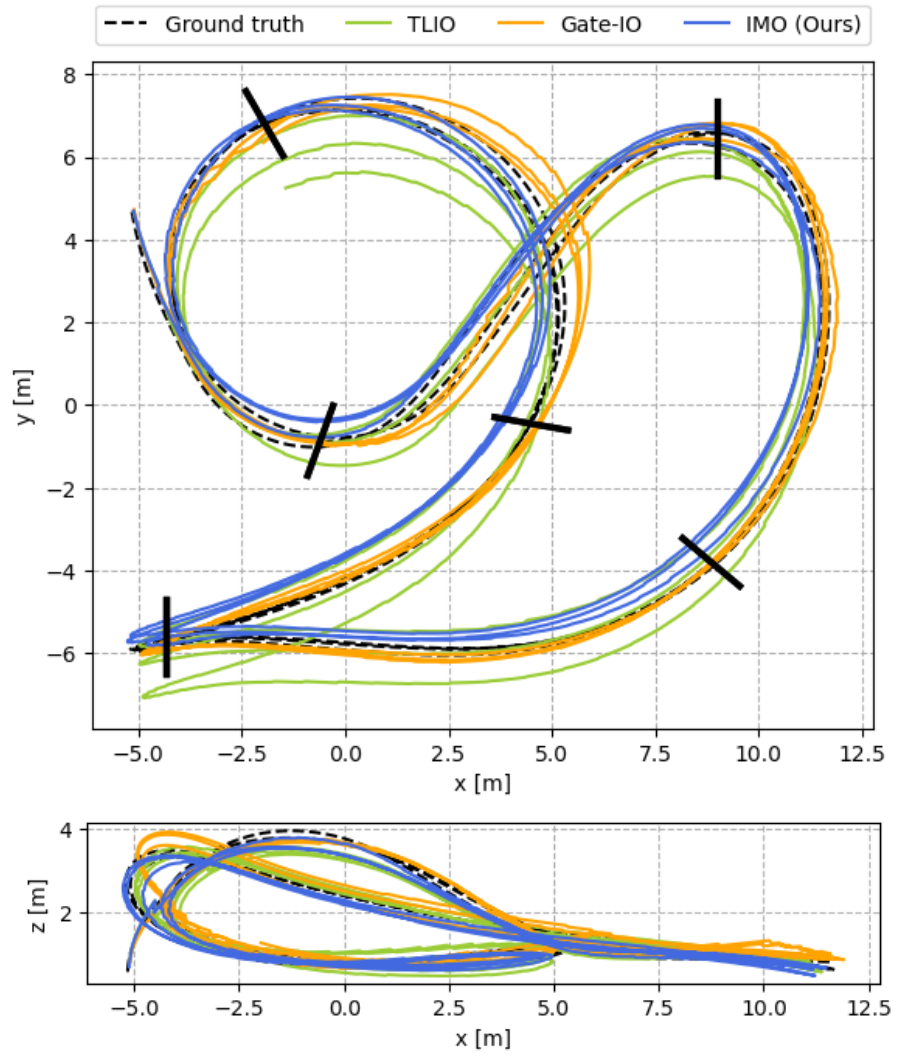}
    \caption{Drone racing evaluation. Trajectory estimated by TLIO, Gate-IO and IMO.}
    \label{fig:splitS_traj}
\end{figure}

\begin{figure}[t!]
    \centering
    \includegraphics[width=0.9\linewidth]{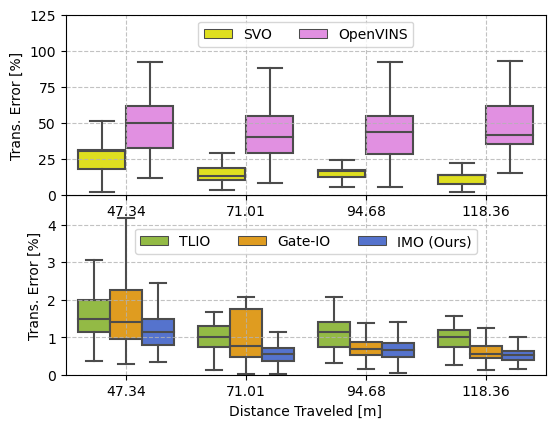}
    \caption{Drone racing evaluation. Relative translation errors achieved by SVO, OpenVINS, TLIO, Gate-IO and IMO.}
    \label{fig:splitS_reltrans_errs}
\end{figure}

\begin{figure}[t!]
    \centering
    \includegraphics[width=0.9\linewidth]{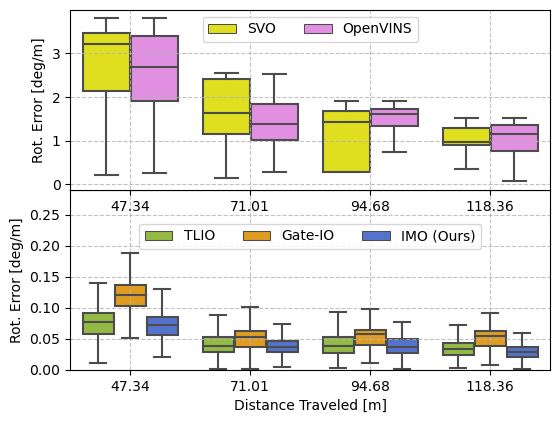}
    \caption{Drone racing evaluation. Relative rotation errors achieved by SVO, OpenVINS, TLIO, Gate-IO, and IMO.}
    \label{fig:splitS_relrot_errs}
\end{figure}

\textit{Evaluation:}
We evaluate the performance of our system in estimating the pose of the quadrotor in a drone racing scenario, cf. Fig.~\ref{fig:eyecatcher}.
The racing track is designed by a professional drone racing pilot and has been used in related works on drone racing~\cite{foehn2021time, romero2022model}.
In each race, the quadrotor flies 3 laps of the track.
In our experiments, the top speed of the autonomous drone is approx. $70\frac{km}{h}$.
\rebuttal{
We use training, validation, and test datasets containing approx. 5, 1.5, and 1.5 min of flight data, respectively. It takes approx. 6 sec to complete a lap of the racing track.}
A visualization of the estimated trajectory by Gate-IO, TLIO, and our algorithm \rebuttal{in a race} is in Fig.~\ref{fig:splitS_traj}.
The two VIO baselines accumulate large drift, cf.~ Table~\ref{tab:absolute_erros}.
We do not include them in Fig.~\ref{fig:splitS_traj} for the sake of clarity.
These results confirm that the classical VIO algorithms fail in drone racing due to perception challenges.
The relative translation and rotation errors \rebuttal{in a race} are shown in Fig.~\ref{fig:splitS_reltrans_errs} and Fig.~\ref{fig:splitS_relrot_errs}, respectively.
The average absolute trajectory errors computed on the test dataset are in Table~\ref{tab:absolute_erros}.
The $\text{ATE}_{\text{T}}$ achieved by our system outperforms TLIO by $54\%$ and is similar to the one achieved by Gate-IO.
\begin{table}[t!]
\caption{
$\text{ATE}_{\text{T}}$ in meters and $\text{ATE}_{\text{R}}$ in degrees of the racing trajectory. In bold is the best value, and in underlined is the second-best value.}
\label{tab:absolute_erros}
\resizebox{0.49\textwidth}{!}{
\begin{tabular}{|c|c|c|c|c|c|c|}
\hline
\multirow{2}{*}{\textbf{\begin{tabular}[c]{@{}l@{}}Trajectory\end{tabular}}} & \multirow{2}{*}{\textbf{\begin{tabular}[c]{@{}c@{}}Eval. \\ metric\end{tabular}}} & \multicolumn{5}{c|}{\textbf{Algorithm}} \\ \cline{3-7} 
 & & \textbf{OpenVINS} & \textbf{SVO} & \textbf{Gate-IO} & \textbf{TLIO} & \textbf{IMO (ours)}\\ \hline \hline 

\multirow{2}{*}{\textit{Racing}} & $\text{ATE}_{\text{T}}$ [m] & 98.20 & 47.20 & \textbf{0.48} & 1.21 & \underline{0.56} \\ & $\text{ATE}_{\text{R}}$ [deg] & 99.00 & 123.00 & \textbf{2.20} & 3.30 & \underline{2.80}\\
\hline
\end{tabular}
}
\end{table}

\subsubsection{Learning the Quadrotor Dynamics}

In this section, we validate the proposed learning-based module, which predicts relative positional displacements from a window of commanded collective thrusts and gyroscope measurements, in two ablation studies.

\begin{figure}[t!]
    \centering
    \includegraphics[width=0.95\linewidth]{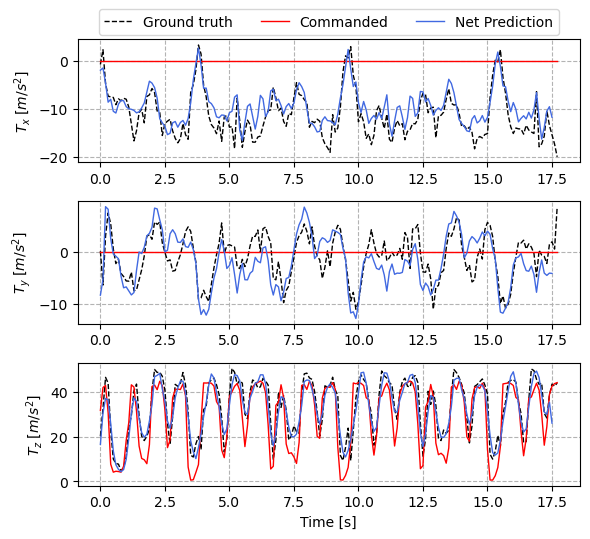}
    \caption{Drone racing evaluation. Comparison of the predictions of the force acting on the quadrotor. \textit{Ground truth} is the force derived from Eq.~\ref{eq:thrust} where the position, velocity, and orientation are from the motion capture system. \textit{Net Prediction} is the force derived from Eq.~\ref{eq:thrust} where the position and velocity are from a 3-DoF trajectory obtained by concatenating the network outputs. The orientation is from the motion caption system.
    \textit{Commanded} is the commanded collective thrust.}
    \label{fig:thrust_comparison}
\end{figure}

In the first ablation study, we analyze the predictions of the force acting on the quadrotor platform.
We compare the predictions of our neural net to the ones obtained by using a motion capture system and to the commanded collective thrust.
The forces predicted by our neural net are derived from Eq.~\ref{eq:thrust} where the position and velocity measurements are obtained from a 3-DoF trajectory that is computed by concatenating the predicted positional displacements of the neural net.
The orientation measurements are from the motion capture system.
Similarly, the forces predicted by the motion capture system are derived from Eq.~\ref{eq:thrust}. 
We show the result of this comparison in Fig.~\ref{fig:thrust_comparison}.
The forces predicted by our neural net match the ones derived from the motion capture system.
We conclude that our neural net learns how to map from the commanded collective thrust to the actual force acting on the quadrotor.
In particular, the neural net also learns to predict the drag force along the body x and y axes, cf. $\text{T}_x$ and $\text{T}_y$ in Fig.~\ref{fig:thrust_comparison}.

\begin{figure}[t!]
    \centering
    \includegraphics[width=0.95\linewidth]{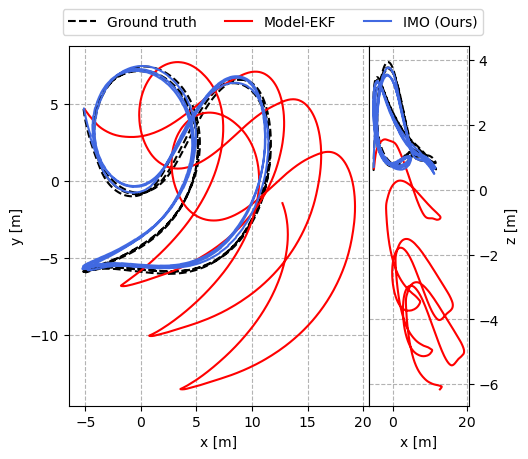}
    \caption{Drone racing evaluation. Comparison against Model-EKF.
    In Model-EKF the measurements $\Delta \mathbf{p}$ are obtained from Eq.~\ref{eq:dp} using the commanded collective thrust as well as orientation and velocity measurements from the motion capture system.}
    \label{fig:modelekf_comparison}
\end{figure}

In the second ablation study, we compare our system against an algorithm that uses an EKF where the measurements $\Delta \mathbf{p}$ are obtained from Eq.~\ref{eq:dp} using the commanded collective thrust and orientation and velocity measurements from the motion capture system.
We call this algorithm Model-EKF.
The propagation step is driven by the IMU measurements as in Eq.~\ref{equ:ekf_propagation}.
The purpose of this study is to show that learning to predict the mismatch between the commanded and applied thrust is essential for accurate state estimation.
A visualization of the estimated trajectory by Model-EKF and the proposed system on one of the test sequences of the racing trajectory is in Fig.~\ref{fig:modelekf_comparison}.
The $\text{ATE}_{\text{T}}$ and $\text{ATE}_{\text{R}}$ achieved by our system are 0.56 m and 2.8 deg, respectively.
The $\text{ATE}_{\text{T}}$ and $\text{ATE}_{\text{R}}$ achieved by Model-EKF are 10.10 m and 4.6 deg, respectively.
These results confirm that the learning-based component of our system is essential to achieve accurate state estimation.

\subsubsection{Filter validation}

\begin{figure}[t!]
    \centering
    \includegraphics[width=0.95\linewidth]{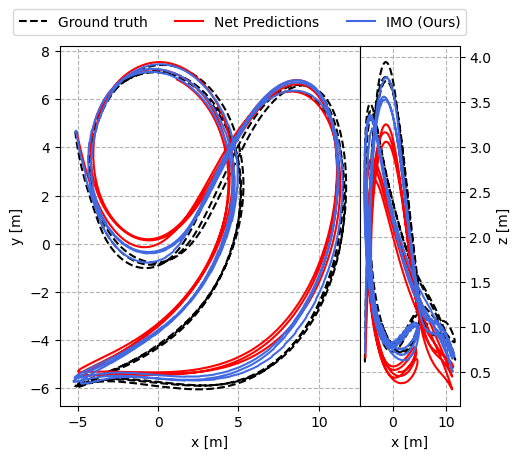}
    \caption{Drone racing evaluation. Comparison of our system against a trajectory computed by concatenating the positional displacements predicted by the neural net.}
    \label{fig:netconc_comparison}
\end{figure}

In this section, we validate the use of the EKF in combination with the neural net.
To this end, we compare the 3-DoF trajectory estimated by our system against a trajectory computed by concatenating the positional displacements predicted by the neural net.
A visualization of the estimated trajectories on one of the test sequences of the racing trajectory is in Fig.~\ref{fig:netconc_comparison}.
The $\text{ATE}_{\text{T}}$ achieved by our system is 0.56 m.
The $\text{ATE}_{\text{T}}$ achieved by concatenating the net predictions is 3.20 m.
This result confirms the superior performance of using the neural net predictions as measurements in an IMU-driven EKF.

\section{Discussion and Conclusion}\label{sec:Conclusions}

In this work, we present a new approach to estimating the state of a quadrotor in autonomous drone racing.
Our method uses a temporal convolutional network to predict the translational motion of the quadrotor using mass-normalized collective thrusts and gyroscope measurements.
The network outputs are relative positional displacements used to update an EKF. 
The EKF is propagated using the IMU measurements.
We have demonstrated that the proposed approach is able to accurately estimate the state of the quadrotor during an autonomous race only relying on an off-the-shelf IMU.
In our experiments, we validate the individual components of our system and we demonstrate that it is superior to the state-of-the-art VIO and IO algorithms in estimating the pose of a racing drone.
Additionally, our system can achieve trajectory estimates similar to those estimated by a VIO that relies on a camera to perform gate detection and has access to the position of the gates.

The main limitation of our approach is that it cannot generalize to trajectories that have not been seen at training time. However, in drone racing competitions, the track is known beforehand. Human pilots spend hours or even days of practice on the race track before the competition. Similarly, our system can be trained with the data collected during practice time and then deployed during the competition. Future work will investigate how to generalize to trajectories that have not been seen at training time. A possible solution is to train the network to estimate the positional displacements in the drone body frame. These displacements can be integrated into a VIO system in order to reduce the dependency on visual inputs. For example, the learned displacement can compensate for informationless visual inputs, e.g. in low-texture scenarios, or low-rate camera measurements.

Although our work focuses specifically on autonomous drone racing, we believe that the proposed approach could have broader implications for reliable state estimation in agile drone flight. In several tasks such as routine inspection and surveillance, the drone is required to fly trajectories that are known beforehand. In these situations, our system can be integrated with a visual-based estimator in order to increase reliability when the visual measurements are degraded, e.g. in low-light conditions.

{\small
\bibliographystyle{IEEEtran}
\bibliography{all}

\begin{thebibliography}{10}
\providecommand{\url}[1]{#1}
\csname url@samestyle\endcsname
\providecommand{\newblock}{\relax}
\providecommand{\bibinfo}[2]{#2}
\providecommand{\BIBentrySTDinterwordspacing}{\spaceskip=0pt\relax}
\providecommand{\BIBentryALTinterwordstretchfactor}{4}
\providecommand{\BIBentryALTinterwordspacing}{\spaceskip=\fontdimen2\font plus
\BIBentryALTinterwordstretchfactor\fontdimen3\font minus
  \fontdimen4\font\relax}
\providecommand{\BIBforeignlanguage}[2]{{%
\expandafter\ifx\csname l@#1\endcsname\relax
\typeout{** WARNING: IEEEtran.bst: No hyphenation pattern has been}%
\typeout{** loaded for the language `#1'. Using the pattern for}%
\typeout{** the default language instead.}%
\else
\language=\csname l@#1\endcsname
\fi
#2}}
\providecommand{\BIBdecl}{\relax}
\BIBdecl

\bibitem{loianno2020special}
G.~Loianno and D.~Scaramuzza, ``Special issue on future challenges and
  opportunities in vision-based drone navigation,'' \emph{J. Field Robot.},
  vol.~37, no.~4, pp. 495--496, 2020.

\bibitem{watkins2020ten}
S.~Watkins, J.~Burry, A.~Mohamed, M.~Marino, S.~Prudden, A.~Fisher, N.~Kloet,
  T.~Jakobi, and R.~Clothier, ``Ten questions concerning the use of drones in
  urban environments,'' \emph{Building and Environment}, vol. 167, p. 106458,
  2020.

\bibitem{rakha2018review}
T.~Rakha and A.~Gorodetsky, ``Review of unmanned aerial system (uas)
  applications in the built environment: Towards automated building inspection
  procedures using drones,'' \emph{Automation in Construction}, vol.~93, pp.
  252--264, 2018.

\bibitem{yang2020online}
Y.~Yang, P.~Geneva, X.~Zuo, and G.~Huang, ``Online imu intrinsic calibration:
  Is it necessary?'' \emph{Proc. of Robotics: Science and Systems (RSS),
  Corvallis, Or}, 2020.

\bibitem{chen2018ionet}
C.~Chen, X.~Lu, A.~Markham, and N.~Trigoni, ``Ionet: Learning to cure the curse
  of drift in inertial odometry,'' in \emph{Proceedings of the AAAI Conference
  on Artificial Intelligence}, vol.~32, no.~1, 2018.

\bibitem{herath2020ronin}
S.~Herath, H.~Yan, and Y.~Furukawa, ``Ronin: Robust neural inertial navigation
  in the wild: Benchmark, evaluations, \& new methods,'' \emph{{IEEE} Int.
  Conf. Robot. Autom. (ICRA)}, pp. 3146--3152, 2020.

\bibitem{liu2020tlio}
W.~Liu, D.~Caruso, E.~Ilg, J.~Dong, A.~I. Mourikis, K.~Daniilidis, V.~Kumar,
  and J.~Engel, ``Tlio: Tight learned inertial odometry,'' \emph{{IEEE} Robot.
  Autom. Lett.}, vol.~5, no.~4, pp. 5653--5660, 2020.

\bibitem{madaan2020airsim}
R.~Madaan, N.~Gyde, S.~Vemprala, M.~Brown, K.~Nagami, T.~Taubner,
  E.~Cristofalo, D.~Scaramuzza, M.~Schwager, and A.~Kapoor, ``Airsim drone
  racing lab,'' in \emph{NeurIPS 2019 Comp. and Demo. Track}, 2020.

\bibitem{foehn2021alphapilot}
P.~Foehn, D.~Brescianini, E.~Kaufmann, T.~Cieslewski, M.~Gehrig, M.~Muglikar,
  and D.~Scaramuzza, ``Alphapilot: Autonomous drone racing,'' \emph{Autonomous
  Robots}, pp. 1--14, 2021.

\bibitem{foehn2021time}
P.~Foehn, A.~Romero, and D.~Scaramuzza, ``Time-optimal planning for quadrotor
  waypoint flight,'' \emph{Science Robotics}, vol.~6, no.~56, 2021.

\bibitem{romero2022model}
A.~Romero, S.~Sun, P.~Foehn, and D.~Scaramuzza, ``Model predictive contouring
  control for time-optimal quadrotor flight,'' \emph{{IEEE} Trans. Robot.},
  2022.

\bibitem{penicka2022minimum}
R.~Penicka and D.~Scaramuzza, ``Minimum-time quadrotor waypoint flight in
  cluttered environments,'' \emph{{IEEE} Robot. Autom. Lett.}, vol.~7, no.~2,
  pp. 5719--5726, 2022.

\bibitem{zhang2019encyclopedia}
D.~Scaramuzza and Z.~Zhang, ``Visual-inertial odometry of aerial robots,''
  \emph{Encyclopedia of Robotics}, 2019.

\bibitem{huang2019visual}
G.~Huang, ``Visual-inertial navigation: A concise review,'' in \emph{{IEEE}
  Int. Conf. Robot. Autom. (ICRA)}.\hskip 1em plus 0.5em minus 0.4em\relax
  IEEE, 2019, pp. 9572--9582.

\bibitem{Antonini18iser}
A.~Antonini, W.~Guerra, V.~Murali, T.~Sayre-McCord, and S.~Karaman, ``The
  blackbird dataset: A large-scale dataset for {UAV} perception in aggressive
  flight,'' in \emph{Int. Symp. Experimental Robotics (ISER)}, 2018.

\bibitem{mourikis2007multi}
A.~I. Mourikis and S.~I. Roumeliotis, ``A multi-state constraint kalman filter
  for vision-aided inertial navigation,'' in \emph{{IEEE} Int. Conf. Robot.
  Autom. (ICRA)}.\hskip 1em plus 0.5em minus 0.4em\relax IEEE, 2007, pp.
  3565--3572.

\bibitem{Leutenegger15ijrr}
S.~Leutenegger, S.~Lynen, M.~Bosse, R.~Siegwart, and P.~Furgale,
  ``Keyframe-based visual-inertial {SLAM} using nonlinear optimization,''
  \emph{Int. J. Robot. Research}, 2015.

\bibitem{brossard2020denoising}
M.~Brossard, S.~Bonnabel, and A.~Barrau, ``Denoising imu gyroscopes with deep
  learning for open-loop attitude estimation,'' \emph{IEEE Robotics and
  Automation Letters}, vol.~5, no.~3, pp. 4796--4803, 2020.

\bibitem{zhang2021imu}
M.~Zhang, M.~Zhang, Y.~Chen, and M.~Li, ``Imu data processing for inertial
  aided navigation: A recurrent neural network based approach,'' \emph{{IEEE}
  Int. Conf. Robot. Autom. (ICRA)}, 2021.

\bibitem{buchanan2021learning}
R.~Buchanan, M.~Camurri, F.~Dellaert, and M.~Fallon, ``Learning inertial
  odometry for dynamic legged robot state estimation,'' \emph{Conf. on Robot.
  Learning (CoRL)}, 2021.

\bibitem{nisar2019vimo}
B.~Nisar, P.~Foehn, D.~Falanga, and D.~Scaramuzza, ``Vimo: Simultaneous visual
  inertial model-based odometry and force estimation,'' \emph{{IEEE} Robot.
  Autom. Lett.}, vol.~4, no.~3, pp. 2785--2792, 2019.

\bibitem{ding2021vid}
Z.~Ding, T.~Yang, K.~Zhang, C.~Xu, and F.~Gao, ``Vid-fusion: Robust
  visual-inertial-dynamics odometry for accurate external force estimation,''
  in \emph{{IEEE} Int. Conf. Robot. Autom. (ICRA)}, 2021, pp. 14\,469--14\,475.

\bibitem{zhang2022dido}
K.~Zhang, C.~Jiang, J.~Li, S.~Yang, T.~Ma, C.~Xu, and F.~Gao, ``Dido: Deep
  inertial quadrotor dynamical odometry,'' \emph{{IEEE} Robot. Autom. Lett.},
  vol.~7, no.~4, pp. 9083--9090, 2022.

\bibitem{svacha2020imu}
J.~Svacha, J.~Paulos, G.~Loianno, and V.~Kumar, ``Imu-based inertia estimation
  for a quadrotor using newton-euler dynamics,'' \emph{IEEE Robotics and
  Automation Letters}, vol.~5, no.~3, pp. 3861--3867, 2020.

\bibitem{oord2016wavenet}
A.~v.~d. Oord, S.~Dieleman, H.~Zen, K.~Simonyan, O.~Vinyals, A.~Graves,
  N.~Kalchbrenner, A.~Senior, and K.~Kavukcuoglu, ``Wavenet: A generative model
  for raw audio,'' \emph{arXiv preprint arXiv:1609.03499}, 2016.

\bibitem{kaufmann2022benchmark}
E.~Kaufmann, L.~Bauersfeld, and D.~Scaramuzza, ``A benchmark comparison of
  learned control policies for agile quadrotor flight,'' \emph{{IEEE} Int.
  Conf. Robot. Autom. (ICRA)}, 2022.

\bibitem{bai2018empirical}
S.~Bai, J.~Z. Kolter, and V.~Koltun, ``An empirical evaluation of generic
  convolutional and recurrent networks for sequence modeling,'' \emph{arXiv
  preprint arXiv:1803.01271}, 2018.

\bibitem{Forster17troSVO}
C.~Forster, Z.~Zhang, M.~Gassner, M.~Werlberger, and D.~Scaramuzza, ``{SVO}:
  Semidirect visual odometry for monocular and multicamera systems,''
  \emph{{IEEE} Trans. Robot.}, vol.~33, no.~2, pp. 249--265, 2017.

\bibitem{geneva2020openvins}
P.~Geneva, K.~Eckenhoff, W.~Lee, Y.~Yang, and G.~Huang, ``Openvins: A research
  platform for visual-inertial estimation,'' in \emph{{IEEE} Int. Conf. Robot.
  Autom. (ICRA)}.\hskip 1em plus 0.5em minus 0.4em\relax IEEE, 2020, pp.
  4666--4672.

\bibitem{Delmerico19icra}
J.~Delmerico, T.~Cieslewski, H.~Rebecq, M.~Faessler, and D.~Scaramuzza, ``Are
  we ready for autonomous drone racing? the {UZH-FPV} drone racing dataset,''
  in \emph{{IEEE} Int. Conf. Robot. Autom. (ICRA)}, 2019.

\bibitem{Zhang18iros}
Z.~Zhang and D.~Scaramuzza, ``A tutorial on quantitative trajectory evaluation
  for visual(-inertial) odometry,'' in \emph{IEEE/RSJ Int. Conf. Intell. Robot.
  Syst. (IROS)}, 2018.

\bibitem{foehn2022agilicious}
P.~Foehn, E.~Kaufmann, A.~Romero, R.~Penicka, S.~Sun, L.~Bauersfeld,
  T.~Laengle, G.~Cioffi, Y.~Song, A.~Loquercio, and D.~Scaramuzza,
  ``Agilicious: Open-source and open-hardware agile quadrotor for vision-based
  flight,'' \emph{Science Robotics}, vol.~7, no.~67, 2022.

\bibitem{song2021autonomous}
Y.~Song, M.~Steinweg, E.~Kaufmann, and D.~Scaramuzza, ``Autonomous drone racing
  with deep reinforcement learning,'' in \emph{2021 IEEE/RSJ International
  Conference on Intelligent Robots and Systems (IROS)}.\hskip 1em plus 0.5em
  minus 0.4em\relax IEEE, 2021, pp. 1205--1212.

\end{thebibliography}
}

\end{document}